\documentclass[sigconf, nonacm]{acmart}
\usepackage{subcaption}

\AtBeginDocument{%
\providecommand\BibTeX{{%
\normalfont B\kern-0.5em{\scshape i\kern-0.25em b}\kern-0.8em\TeX}}}

\copyrightyear{2021}
\acmYear{2021}
\setcopyright{acmcopyright}
\acmConference[MMAsia '21]{ACM Multimedia Asia}{December 1--3, 2021}{Gold Coast, Australia}
\acmBooktitle{ACM Multimedia Asia (MMAsia '21), December 1--3, 2021, Gold Coast, Australia}
\acmPrice{15.00}
\acmDOI{10.1145/3469877.3490571}
\acmISBN{978-1-4503-8607-4/21/12}

\begin{document}

\title[Utilizing Resource-Rich Language Datasets for End-to-End Scene Text Recognition in Resource-Poor Languages]{Utilizing Resource-Rich Language Datasets for End-to-End \\ Scene Text Recognition in Resource-Poor Languages}

\author{Shota Orihashi, Yoshihiro Yamazaki, Naoki Makishima, Mana Ihori, \linebreak Akihiko Takashima, Tomohiro Tanaka, and Ryo Masumura}
\affiliation{%
\institution{NTT Computer and Data Science Laboratories, NTT Corporation, Japan}
\country{\vspace{\baselineskip}}
}

\renewcommand{\shortauthors}{S. Orihashi et al.}

\begin{abstract}
This paper presents a novel training method for end-to-end scene text recognition. End-to-end scene text recognition offers high recognition accuracy, especially when using the encoder-decoder model based on Transformer. To train a highly accurate end-to-end model, we need to prepare a large image-to-text paired dataset for the target language. However, it is difficult to collect this data, especially for resource-poor languages. To overcome this difficulty, our proposed method utilizes well-prepared large datasets in resource-rich languages such as English, to train the resource-poor encoder-decoder model. Our key idea is to build a model in which the encoder reflects knowledge of multiple languages while the decoder specializes in knowledge of just the resource-poor language. To this end, the proposed method pre-trains the encoder by using a multilingual dataset that combines the resource-poor language's dataset and the resource-rich language's dataset to learn language-invariant knowledge for scene text recognition. The proposed method also pre-trains the decoder by using the resource-poor language's dataset to make the decoder better suited to the resource-poor language. Experiments on Japanese scene text recognition using a small, publicly available dataset demonstrate the effectiveness of the proposed method.
\end{abstract}




\maketitle

\newcommand{\bm}[1]{{\mbox{\boldmath $#1$}}}
\newcommand{\argmin}{\mathop{\rm arg~min}\limits}

\section{Introduction}
Natural scene images contain a lot of textual information such as store advertising signs and traffic signs. Scene text recognition is the task of identifying texts present in images such as signs detected from natural scene images. Scene text recognition can be applied to many tasks such as image classification \cite{Karaoglu17a, Karaoglu17b, Bai18}, image retrieval \cite{Karaoglu17a, Gomez18}, and visual question answering \cite{Biten19}. Due to the appeal of these various potential applications, research and development of scene text recognition technology is being actively carried out in academic and industrial fields.

With improvements in deep learning technology, many scene text recognition methods have been proposed \cite{Long21}. Most of the conventional methods are based on end-to-end neural networks. Typically, the input image is converted into continuous representations by using convolutional neural networks (CNNs) and bidirectional long short-term memory recurrent neural networks (BLSTM-RNNs). The obtained continuous representations are then subjected to connectionist temporal classification (CTC) \cite{Graves06} which yields a character string \cite{Shi16end, Liu16, Wang17, Borisyuk18}. Other methods use encoder-decoder models that utilize LSTM and attention mechanisms instead of CTC for sequence-to-sequence conversion \cite{Shi16robust, Lee16, Baek19}. Also, in recent years, due to improvements in deep learning technology for natural language processing, high recognition accuracy is possible from encoder-decoder models that utilize Transformer \cite{Vaswani17}; sequence-to-sequence conversion is realized solely by attention mechanisms, not by RNNs such as LSTM \cite{Sheng19, Zhu19, Wang19, Yu20, Bleeker20, Lu21}. These methods extract continuous representations, which capture the features of the input image, in the encoder by using CNN and Transformer encoder, while the decoder translates the continuous representations into character strings by using Transformer decoder.

To train a highly accurate end-to-end encoder-decoder scene text recognition model, a large image-to-text paired training dataset in the target language is required. While various well-prepared large public datasets are available for English \cite{Mishra12, Jaderberg14, Gupta16}, there are few public datasets for minor languages such as Japanese. In fact, the public Japanese scene text dataset provided for the {\it ICDAR2019 robust reading challenge on multilingual scene text detection and recognition} \cite{Nayef19} has a relatively small amount of data in Japanese (10K order) compared to publicly available English datasets (1M order). Of course, several methods have been proposed to synthesize training data \cite{Gupta16, Zhan18, Long20}, but these methods require the preparation of an appropriate corpus and images in the target language and the target domain in advance. These preparations also require expertise and are costly. Therefore, we need a training method that can yield an accurate scene text recognition model for the target language from small datasets in the target language.

In this paper, we present a novel training method for an encoder-decoder scene text recognition model for resource-poor languages. Our method utilizes well-prepared large datasets in resource-rich languages such as English to train a scene text recognition model for resource-poor target languages. Our key idea is to build a model in which the encoder reflects the knowledge available in multiple languages for scene text recognition including a variety of background images and character string shapes such as curved and tilted, and the decoder specializes in just the knowledge of the resource-poor language. To this end, the proposed method pre-trains the encoder of the model by using a multilingual dataset, a combination of the resource-poor language's dataset and the resource-rich language's dataset, to learn language-invariant knowledge for scene text recognition. Our method also pre-trains the decoder of the model by the resource-poor language's dataset to ensure that the decoder is specific for the resource-poor language. The proposed method finally fine-tunes the pre-trained encoder and decoder in the resource-poor language. Our method enables us to train the model efficiently without a large dataset in the resource-poor target language. Experiments on a small publicly available Japanese dataset \cite{Nayef19} and a large English dataset \cite{Jaderberg14} demonstrate the effectiveness of the proposed method.

Our contributions are summarized as follows:
\begin{itemize}
\item We provide a training method for an end-to-end encoder-decoder scene text recognition model for resource-poor languages that utilizes well-prepared large datasets in resource-rich languages effectively. To the best of our knowledge, while training methods utilizing multilingual data for end-to-end models have been proposed in the fields of speech and language processing \cite{Adams19, Conneau19, Liu20}, ours is the first work to utilize multilingual data in training a scene text recognition model.
\item We conduct experiments on Japanese scene text recognition using highly accurate Transformer-based scene text recognition models \cite{Sheng19, Wang19} with a detailed ablation study that verifies the effectiveness of the proposed approach. The experiments show that even a small amount of resource-rich language's data improves performance in the resource-poor language.
\end{itemize}

\section{Transformer-Based Scene Text Recognition}

This section describes scene text recognition based on Transformer \cite{Sheng19, Wang19}. Transformer \cite{Vaswani17} was originally proposed for machine translation and is based solely on attention mechanisms; it has been successful in various natural language processing tasks. In recent years, inspired by the machine translation model, scene text recognition methods based on Transformer have been proposed \cite{Sheng19, Zhu19, Wang19, Yu20, Bleeker20, Lu21}. High recognition accuracy has been obtained due to its powerful language modeling abilities.

Scene text recognition is a task that estimates character string $\bm{C}=\{c_1,\cdots,c_T\}$ from character image $\bm{I}$, where $c_t$ is the $t$-th character of the string, and $T$ is the number of characters. In the auto-regressive encoder-decoder recognition model based on Transformer, the generation probability of $\bm{C}$ from $\bm{I}$ is modeled as
\begin{equation}
P(\bm{C}\mid\bm{I};\bm{\Theta})=\prod_{t=1}^{T}P(c_t\mid c_{1:t-1}, \bm{I};\bm{\Theta}),
\end{equation}
where $c_{1:t-1}=\{c_1,\cdots,c_{t-1}\}$, and $\bm{\Theta}=\{\bm{\theta}_{\rm enc}, \bm{\theta}_{\rm dec}\}$ represents the trainable model parameter set.

Figure 1 shows network structures for a scene text recognition model based on Transformer \cite{Sheng19, Wang19}. As shown in Figure 1, the scene text recognition model based on Transformer consists of an encoder and a decoder.

\begin{figure*}[t]
\begin{minipage}[b]{0.48\linewidth}
\centering
\includegraphics[width=0.95\linewidth]{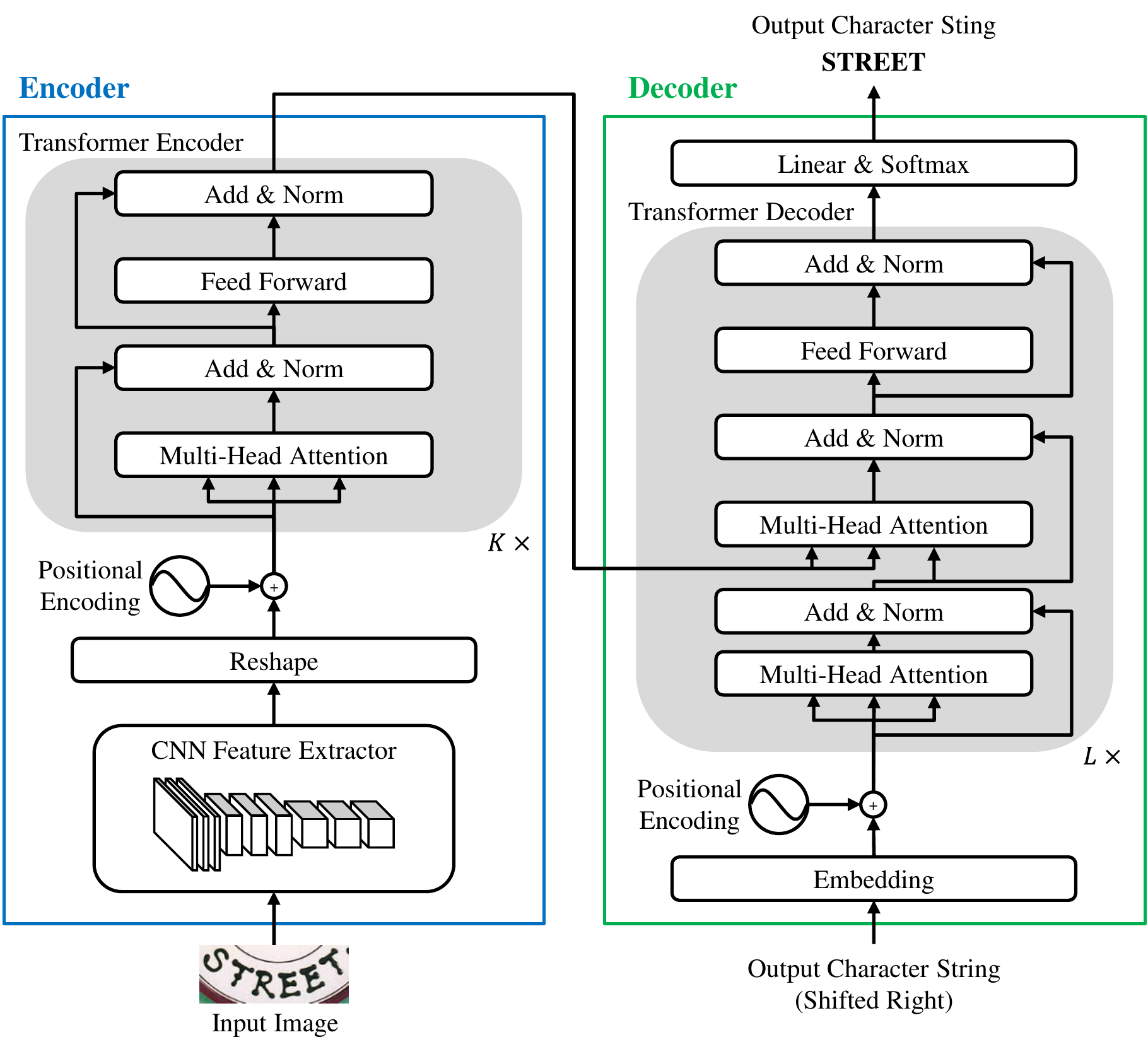}
\subcaption{modeling by Sheng \textit{et al.} \cite{Sheng19}}
\end{minipage}
\begin{minipage}[b]{0.48\linewidth}
\centering
\includegraphics[width=0.95\linewidth]{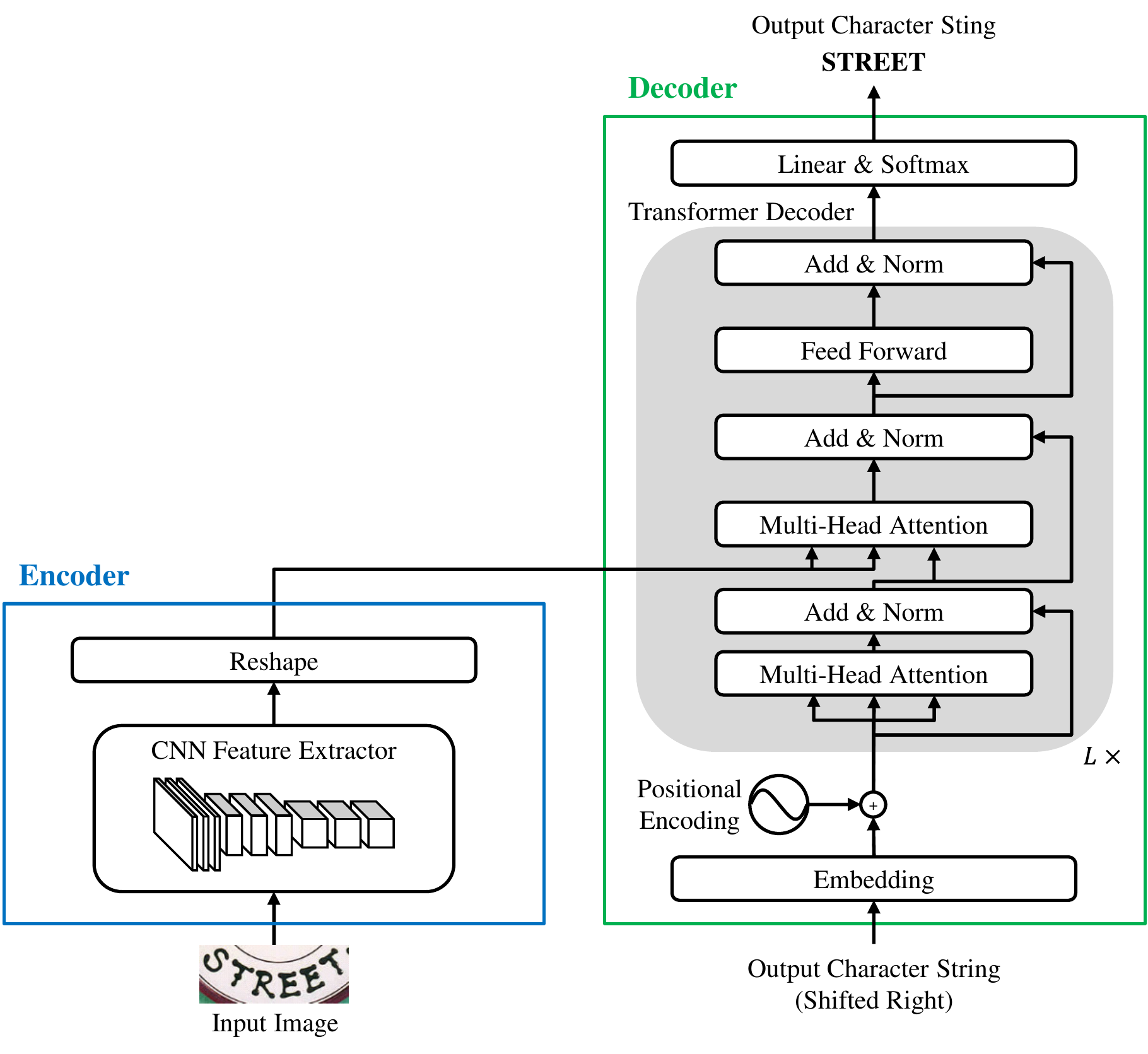}
\subcaption{modeling by Wang \textit{et al.} \cite{Wang19}}
\end{minipage}
\caption{Network structures of a Transformer-based encoder-decoder model for scene text recognition.}
\label{fig:network}
\end{figure*}

\subsection{Encoder}

In the encoder, input image $\bm{I}$ is converted into continuous vectors $\bm{Q}=\{\bm{q}_1,\cdots,\bm{q}_J\}$ as 
\begin{equation}
\bm{P}={\rm CNNFeatureExtractor}(\bm{I};\bm{\theta}_{\rm enc}),
\end{equation}
\begin{equation}
\bm{Q}={\rm Reshape}(\bm{P}),
\end{equation}
where ${\rm CNNFeatureExtractor}()$ is a function that extracts image features by using a CNN-based model, ${\rm Reshape}()$ is a function that translates three-dimensional image features (${\rm width}\times {\rm height} \times {\rm channels}$) into two-dimensional vectors (${\rm width}\times ({\rm height}\times{\rm channels})$), and $J$ is the width of the continuous vectors. The encoder proposed by Wang {\it et al.} \cite{Wang19} outputs continuous vectors $\bm{Q}$, see Figure 1 (b). On the other hand, in the encoder proposed by Sheng {\it et al.} \cite{Sheng19} shown in Figure 1 (a), continuous vectors $\bm{Q}$ are then projected into $\bm{R}=\{\bm{r}_1,\cdots,\bm{r}_J\}$ for input to the Transformer encoder block as
\begin{equation}
\bm{r}_j={\rm AddPosEnc}(\bm{q}_j),
\end{equation}
where ${\rm AddPosEnc}()$ is a function that adds a continuous vector in which position information is embedded. The Transformer encoder composes continuous vectors $\bm{S}^{(K)}$ from $\bm{R}$ by using $K$ Transformer encoder blocks. The $k$-th Transformer encoder block forms the $k$-th continuous vectors $\bm{S}^{(k)}$ from the lower layer inputs $\bm{S}^{(k-1)}$ as
\begin{equation}
\bm{S}^{(k)}={\rm TransformerEnc}(\bm{S}^{(k-1)};\bm{\theta}_{\rm enc}),
\end{equation}
where $\bm{S}^{(0)}=\bm{R}$, and ${\rm TransformerEnc}()$ is a Transformer encoder block that consists of a scaled dot product multi-head self-attention layer and a position-wise feed-forward network \cite{Vaswani17}. The encoder proposed by Sheng {\it et al.} \cite{Sheng19} outputs continuous vectors $\bm{S}^{(K)}$.

\subsection{Decoder}

The decoder computes the generation probability of a character string from the preceding character string and continuous vectors output from the encoder. The predicted probabilities of the $t$-th character, $c_t$, are calculated as
\begin{equation}
P(c_t\mid c_{1:t-1},\bm{I};\bm{\Theta})={\rm Softmax}(\bm{u}_{t-1}^{(L)};\bm{\theta}_{\rm dec}),
\end{equation}
where ${\rm Softmax}()$ is a softmax layer with linear transformation. The Transformer decoder forms hidden representation $\bm{u}_{t-1}^{(L)}$ from encoder output $\bm{V}$ by using $L$ Transformer decoder blocks, where
\begin{equation}
\bm{V}=
\begin{cases}
\bm{S}^{(K)} & \text{in modeling by Sheng {\it et al.} \cite{Sheng19}}, \\
\bm{Q} & \text{in modeling by Wang {\it et al.} \cite{Wang19}}.
\end{cases}
\end{equation}
The $l$-th Transformer decoder block forms the $l$-th hidden representation $\bm{u}_{t-1}^{(l)}$ from the lower layer inputs $\bm{U}_{1:t-1}^{(l-1)}=\{\bm{u}_{1}^{(l-1)},\cdots,\bm{u}_{t-1}^{(l-1)}\}$ as
\begin{equation}
\bm{u}_{t-1}^{(l)}={\rm TransformerDec}(\bm{U}_{1:t-1}^{(l-1)},\bm{V};\bm{\theta}_{\rm dec}),
\end{equation}
where ${\rm TransformerDec}()$ is a Transformer decoder block that consists of a scaled dot product multi-head self-attention layer, a scaled dot product multi-head source-target attention layer, and a position-wise feed-forward network \cite{Vaswani17}. The hidden representation $\bm{u}_{t-1}^{(0)}$ is given by
\begin{equation}
\bm{u}_{t-1}^{(0)}={\rm AddPosEnc}(\bm{c}_{t-1}),
\end{equation}
\begin{equation}
\bm{c}_{t-1}={\rm Embedding}(c_{t-1};\bm{\theta}_{\rm dec}),
\end{equation}
where ${\rm Embedding}()$ is a linear layer that embeds the input character into a continuous vector.

\subsection{Training}

To train the model, the target language's dataset $\mathcal{D}=\{(\bm{I}_1,\bm{C}_1),\cdots,\\(\bm{I}_N,\bm{C}_N)\}$ is used, where $N$ is the number of data points. The optimization of model parameters is represented as
\begin{equation}
\hat{\bm{\theta}}_{\rm enc}, \hat{\bm{\rm \theta}}_{\rm dec}=\argmin_\bm{\Theta}\sum_{n=1}^N\bigl\{-\log{P(\bm{C}_n\mid\bm{I}_n;\bm{\Theta})}\bigr\},
\end{equation}
where $\hat{\bm{\theta}}_{\rm enc}$ and $\hat{\bm{\theta}}_{\rm dec}$ are trained parameters.

In this paper, to train an accurate scene text recognition model for the target language when a large dataset in the target language is not available, i.e. the number of data points, $N$, for the target language is small, we propose a training method that utilize not only the dataset in the resource-poor target languages but also a well-prepared dataset of a resource-rich language.

\section{Proposed Method}

This section details the proposed method. The proposed method trains the encoder-decoder scene text recognition model based on Transformer described in Section 2 for the resource-poor target language.

The main idea of the proposed method is to utilize a well-prepared dataset of a resource-rich language such as English to train the recognition model for the resource-poor language. The proposed method does not simply train the model on a dataset composed of two languages, but trains the resource-poor language's model by using efficient combinations of the resource-poor language's dataset and the resource-rich language's dataset. In detail, the proposed method optimizes two models in pre-training at first, and then a part of each model is used as a pre-trained encoder and a pre-trained decoder for fine-tuning. For encoder pre-training, the proposed method utilizes the multilingual dataset composed of the resource-poor language's dataset and the resource-rich language's dataset. By this approach, the encoder can be trained by using a larger dataset than that possible when using only the resource-poor language's dataset. Therefore, the encoder can learn features of images beyond languages including a variety of images and character string shapes such as curved and tilted from the larger dataset, which improves robustness effectively. On the other hand, for decoder pre-training, the proposed method utilizes only the resource-poor language's dataset. This approach specializes the decoder for the resource-poor language. Since the decoder translates the image features captured by the encoder into character strings, recognition accuracy can be improved by specializing the decoder for the resource-poor language. Finally, the pre-trained encoder and decoder are fine-tuned by using the resource-poor language's dataset.

Figure 2 outlines the proposed method. The proposed method consists of two training phases; pre-training and fine-tuning.

\begin{figure*}[t]
\centering
\includegraphics[width=0.95\linewidth]{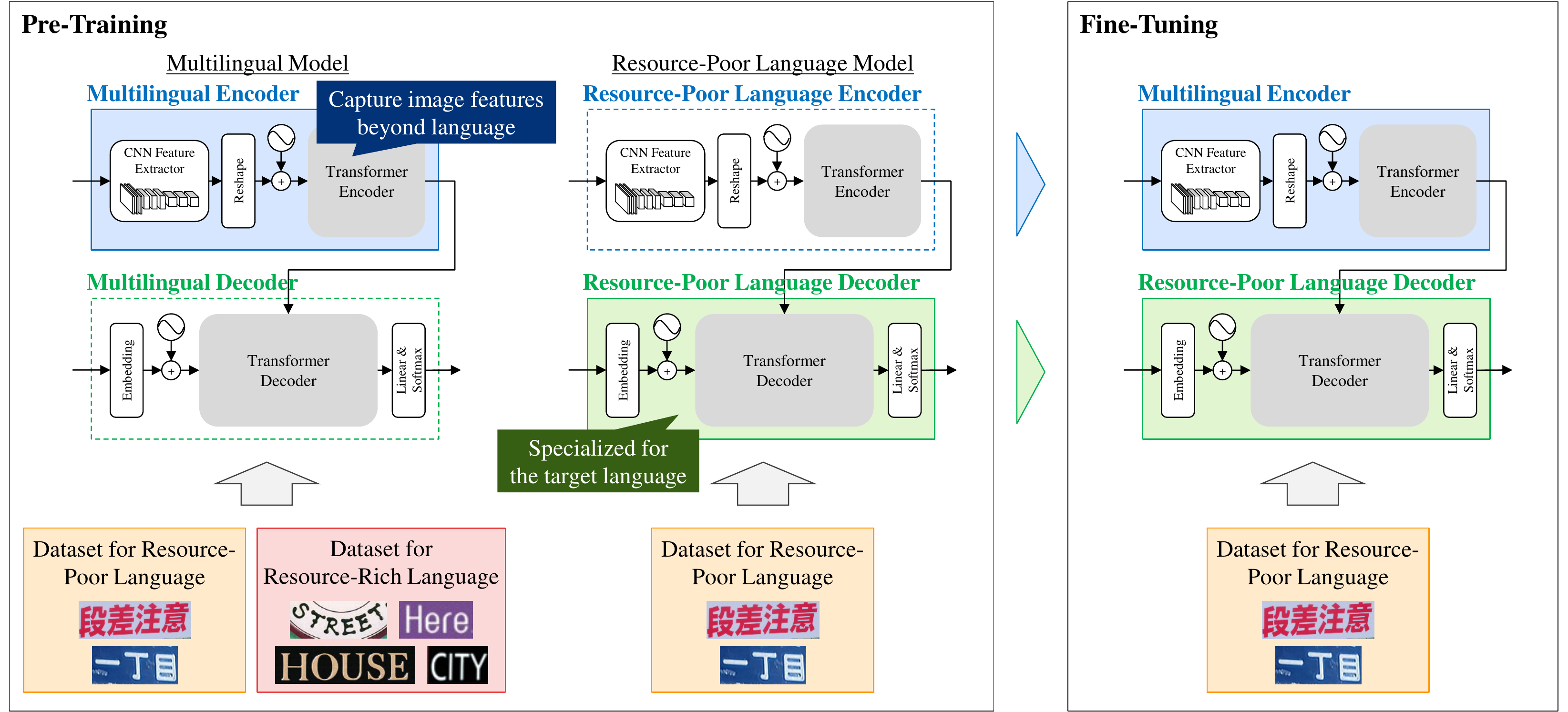}
\caption{Outline of the proposed method.}
\label{fig:method}
\end{figure*}

\subsection{Pre-Training}

In the pre-training phase, the proposed method uses not only the resource-poor language's dataset $\mathcal{D}^{\rm rp}=\{(\bm{I}^{\rm rp}_1,\bm{C}^{\rm rp}_1),\cdots,(\bm{I}^{\rm rp}_N,\bm{C}^{\rm rp}_N)\}$, but also a well-prepared large dataset in a resource-rich language $\mathcal{D}^{\rm rr}=\{(\bm{I}^{\rm rr}_1,\bm{C}^{\rm rr}_1),\cdots,(\bm{I}^{\rm rr}_M,\bm{C}^{\rm rr}_M)\}$, where $N$ and $M$ are the number of data points in the resource-poor language's dataset and the resource-rich language's dataset, respectively. We assume $N<M$. The proposed method pre-trains the encoder and the decoder by using the multilingual dataset and the resource-poor language's dataset, respectively.

As for encoder pre-training, the multilingual model that consists of multilingual encoder (ME) and multilingual decoder (MD) is trained. Training uses a multilingual dataset made by combining the resource-poor language's dataset $\mathcal{D}^{\rm rp}$ with the resource-rich language's dataset $\mathcal{D}^{\rm rr}$. The optimization of model parameters is given by
\begin{equation}
\begin{split}
\dot{\bm{\theta}}_{\rm enc}, \dot{\bm{\theta}}_{\rm dec}=\argmin_\bm{\Theta}\left[\sum_{n=1}^{N}\bigl\{-\log{P(\bm{C}^{\rm rp}_n\mid\bm{I}^{\rm rp}_n;\bm{\Theta})\bigr\}}\right.\\
\left.+\sum_{m=1}^{M}\bigl\{-\log{P(\bm{C}^{\rm rr}_m\mid\bm{I}^{\rm rr}_m;\bm{\Theta})}\bigr\}\right],
\end{split}
\end{equation}
where $\dot{\bm{\theta}}_{\rm enc}$ and $\dot{\bm{\theta}}_{\rm dec}$ are the trained parameters of ME and MD in encoder pre-training.

As for decoder pre-training, the resource-poor language model that consists of resource-poor language encoder (RPLE) and resource-poor language decoder (RPLD) is trained. Training uses the resource-poor language's dataset $\mathcal{D}^{\rm rp}$. The optimization of model parameters is given by
\begin{equation}
\ddot{\bm{\theta}}_{\rm enc}, \ddot{\bm{\theta}}_{\rm dec}=\argmin_\bm{\Theta}\sum_{n=1}^{N}\bigl\{-\log{P(\bm{C}^{\rm rp}_n\mid\bm{I}^{\rm rp}_n;\bm{\Theta})\bigr\}},
\end{equation}
where $\ddot{\bm{\theta}}_{\rm enc}$ and $\ddot{\bm{\theta}}_{\rm dec}$ are the trained parameters RPLE and RPLD in decoder pre-training. 

\subsection{Fine-Tuning}

In the fine-tuning phase, the proposed method trains the final recognition model using the pre-trained encoder-decoder parameters. Thus, parameters of ME $\dot{\bm{\theta}}_{\rm enc}$ pre-trained by encoder pre-training and RPLD $\ddot{\bm{\theta}}_{\rm dec}$ pre-trained by decoder pre-training are used as the initial values for fine-tuning. Fine-tuning is carried out by using the resource-poor language's dataset $\mathcal{D}^{\rm rp}$. The optimization of model parameters is given by
\begin{equation}
\hat{\dot{\bm{\theta}}}_{\rm enc}, \hat{\ddot{\bm{\theta}}}_{\rm dec}=\argmin_\bm{\Theta}\sum_{n=1}^{N}\bigl\{-\log{P(\bm{C}^{\rm rp}_n\mid\bm{I}^{\rm rp}_n;\bm{\Theta})\bigr\}},
\end{equation}
where $\hat{\dot{\bm{\theta}}}_{\rm enc}$ and $\hat{\ddot{\bm{\theta}}}_{\rm dec}$ are the fine-tuned final parameters.

In the experiments in Section 4, we additionally examined a training procedure that fine-tuned the parameters of ME $\dot{\bm{\theta}}_{\rm enc}$ with a randomly initialized decoder, and a training procedure that fine-tuned the parameters of ME $\dot{\bm{\theta}}_{\rm enc}$ and MD $\dot{\bm{\theta}}_{\rm dec}$.

\section{Experiment}

We conducted experiments to confirm the effectiveness of the proposed method. We selected Japanese as the resource-poor target language, for which no public large dataset is available. In addition to the limitation posed by the small available dataset, Japanese scene text recognition is further complicated by the diversity of characters. We selected English as the resource-rich language.

\subsection{Datasets}

We used the Japanese dataset created for the {\it ICDAR2019 robust reading challenge on multilingual scene text detection and recognition} \cite{Nayef19}; it is the only publicly available Japanese scene text dataset, for the resource-poor language's dataset. The ICDAR2019 dataset holds annotated real and synthesized image data created using the synthesizing method of \cite{Gupta16} for end-to-end scene text detection and recognition in 10 languages. To construct the Japanese scene text recognition dataset for this experiment, we first selected Japanese real image data and synthesized image data. Then, we cropped the images according to the annotation for scene text detection, and excluded data that contained characters other than standard Japanese characters. This yielded 9,346 real images and 65,452 synthesized images. Mixing and splitting these images at the ratio of $9:1$ yielded training data of 67,368 images and test data of 7,430 images. The training data was used as $\mathcal{D}^{\rm rp}$ for both pre-training and fine-tuning. There were 2,332 character classes.

We used MJSynth \cite{Jaderberg14} as the resource-rich language's (i.e. English) dataset, $\mathcal{D}^{\rm rr}$. The number of training data images was 8,027,346.

\begin{table*}[t]
\caption{Measured results in terms of recognition accuracy.}
\label{tab:accuracy}
\begin{tabular}{llllr}
\toprule
Modeling & Training Procedure & Encoder Pre-Training & Decoder Pre-Training & Accuracy (\%) \\
\midrule
Shi {\it et al.} \cite{Shi16end} & Baseline & -- & -- & 27.82 \\
Borisyuk {\it et al.} \cite{Borisyuk18} & Baseline & -- & -- & 29.73 \\
Shi {\it et al.} \cite{Shi16robust} & Baseline & -- & -- & 30.70 \\
Lee {\it et al.} \cite{Lee16} & Baseline & -- & -- & 47.85 \\
Wang {\it et al.} \cite{Wang17} & Baseline & -- & -- & 48.05 \\
Liu {\it et al.} \cite{Liu16} & Baseline & -- & -- & 54.48 \\
Baek {\it et al.} \cite{Baek19} & Baseline & -- & -- & 55.34 \\
\midrule
Sheng {\it et al.} \cite{Sheng19} & Baseline & -- & -- & 47.85 \\
Sheng {\it et al.} \cite{Sheng19} & Training w/ ME & Multilingual & -- & 59.30 \\
Sheng {\it et al.} \cite{Sheng19} & Training w/ ME+MD & Multilingual & Multilingual & 59.43 \\
Sheng {\it et al.} \cite{Sheng19} & Proposed training w/ ME+RPLD & Multilingual & Japanese & \textbf{62.14} \\
\midrule
Wang {\it et al.} \cite{Wang19} & Baseline & -- & -- & 65.22 \\
Wang {\it et al.} \cite{Wang19} & Training w/ ME & Multilingual & -- & 69.58 \\
Wang {\it et al.} \cite{Wang19} & Training w/ ME+MD & Multilingual & Multilingual & 69.33 \\
Wang {\it et al.} \cite{Wang19} & Proposed training w/ ME+RPLD & Multilingual & Japanese & \textbf{72.57} \\
\bottomrule
\end{tabular}
\end{table*}

\begin{figure}[t]
\centering
\includegraphics[width=0.95\linewidth]{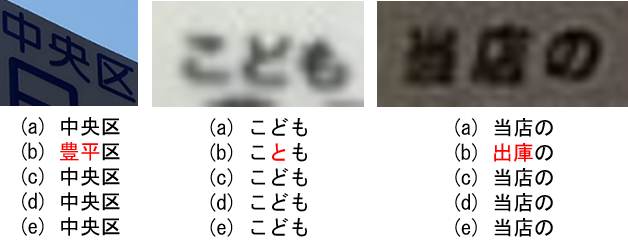}
\caption{Example of recognition results improved by pre-training encoder utilizing multilingual dataset.}
\label{fig:example}
\end{figure}

\subsection{Setups}

We tested the following four training procedures to evaluate the proposed training method. \textbf{Baseline} did not pre-train the recognition model; training used only the Japanese dataset from scratch. In \textbf{Training w/ ME}, we pre-trained ME by using the multilingual dataset made by combining the Japanese dataset and the English dataset, and fine-tuned ME with a randomly initialized decoder by using the Japanese dataset. In \textbf{Training w/ ME+MD}, we pre-trained ME and MD by using the multilingual dataset, and fine-tuned them by using the Japanese dataset. In \textbf{Proposed training w/ ME+RPLD}, we pre-trained ME and RPLD by using the multilingual dataset and the Japanese dataset respectively, and fine-tuned them by using the Japanese dataset.

Two recognition models based on Transformer were evaluated. The first one is the model proposed by Sheng {\it et al.} \cite{Sheng19}; VGG16 \cite{Simonyan15} up to the 10-th convolution layer was applied as the CNN feature extractor. The second one is the model proposed by Wang {\it et al.} \cite{Wang19}; ResNet34 \cite{He16} was applied as the CNN feature extractor. The Transformer blocks were composed under the following conditions: the number of Transformer encoder blocks and Transformer decoder blocks were set to 1, the dimensions of the output continuous representations and the inner outputs in the position-wise feed forward networks were set to 512, and the number of heads in the multi-head attentions was set to 4. We also evaluated the models based on RNNs \cite{Shi16end, Liu16, Wang17, Borisyuk18, Shi16robust, Lee16, Baek19}. For the models based on RNNs, we evaluated only the baseline training procedure, and the model structures followed the evaluation by \cite{Baek19}.

For all evaluated models, the input image size was $400 \times 64$, and the input images were scaled to satisfy the resolution. For training, the optimizer was based on stochastic gradient descent (SGD) with learning rate of 0.01. The mini-batch size was set to 32 images. Note that a part of the training data was used for early stopping. As an evaluation metric, we used the recognition accuracy as determined by exact matching.

\begin{figure*}[t]
\centering
\includegraphics[width=0.90\linewidth]{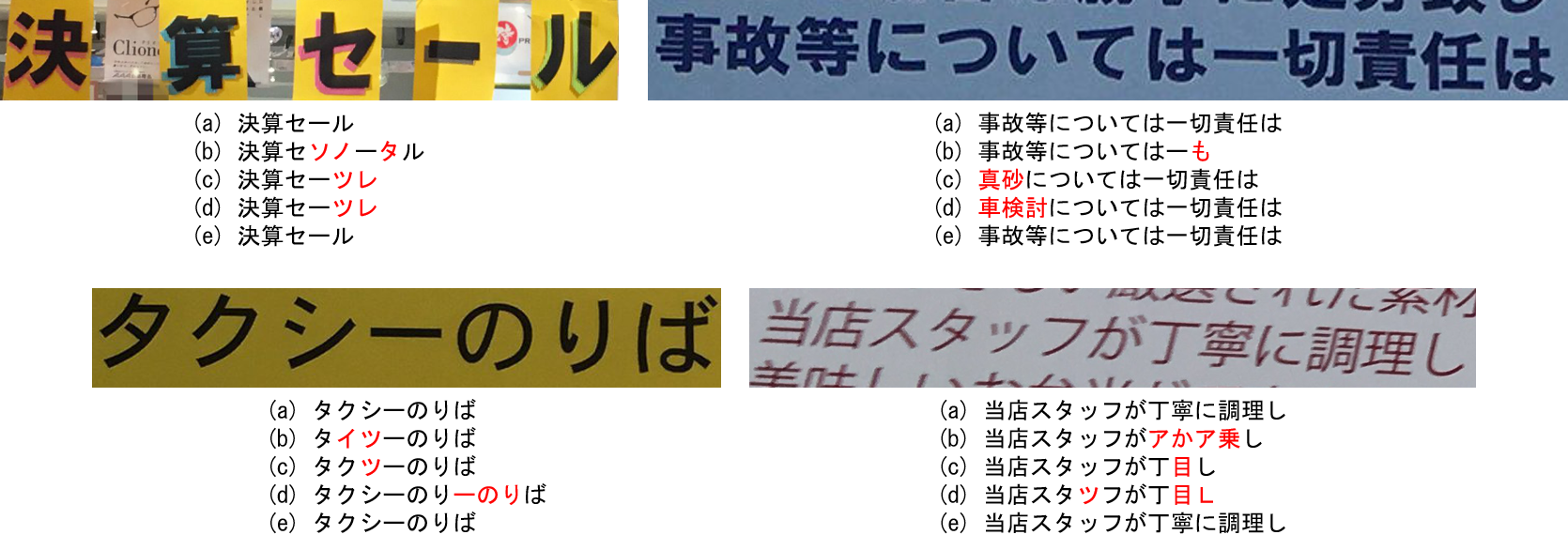}
\caption{Example of recognition results improved by pre-training decoder utilizing Japanese dataset.}
\label{fig:example}
\end{figure*}

\subsection{Results}

The recognition accuracy values are shown in Table 1. First, the recognition models based on Transformer have the same or higher Japanese character recognition accuracy than the recognition model based on RNNs with limited training data. Next, for models based on Transformer, the accuracy obtained by baseline was low compared to the use of pre-training, and utilizing pre-training improved the accuracy. In detail, when we utilized training with ME or ME+MD, the performance improvement was small. On the other hand, our training proposal utilizing ME+RPLD further improved the recognition accuracy. Examples of characters extracted by the model of Wang {\it et al.} \cite{Wang19} are shown in Figures 3 and 4. Note that the captions for each image on Figures 3 and 4 are (a) ground truth, recognition results of (b) baseline, (c) training w/ ME, (d) training w/ ME+MD, and (e) proposed training w/ ME+RPLD. The results in Figure 3 show that the proposed training method can prevent the erroneous recognition of words in tilted or blurred images; this is mainly the effect of utilizing the multilingual dataset in pre-training the encoder. The results in Figure 4 show that the proposed training method can prevent the false recognition of words that do not exist given relatively long character strings, which is mainly due to pre-training the decoder by utilizing the resource-poor language's dataset. These results confirm that the proposed training method is an effective way of improving recognition accuracy when the image-to-text paired dataset in the resource-poor language is limited.

We also evaluated the recognition accuracy while varying the size of the English dataset in the pre-training phase. We prepared English datasets of four sizes: 8M as used in the above evaluation, 4M (4,013,673 data), 2M (2,006,837 data), and 1M (1,076,675 data). The resulting accuracy values are shown in Table 2. They show that increasing the data size improves the recognition accuracy. On the other hand, the proposed method is more effective than baseline even when the size of the English dataset is small.

\begin{table}[t]
\caption{Measured results of proposed training in terms of recognition accuracy (\%) versus size of English dataset.}
\label{tab:result}
\begin{tabular}{lrr}
\toprule
Data Size & Sheng {\it et al.} \cite{Sheng19} & Wang {\it et al.} \cite{Wang19} \\
\midrule
0 (Same as baseline) & 47.85 & 65.22 \\
1M & 57.90 & 68.52 \\
2M & 59.07 & 70.03 \\
4M & 59.69 & 71.28 \\
8M & \textbf{62.14} & \textbf{72.57} \\
\bottomrule
\end{tabular}
\end{table}

\section{Conclusion}

This paper proposed a novel training method for an encoder-decoder scene text recognition model for resource-poor languages. The key advance of our method is to utilize a well-prepared large dataset in a resource-rich language, and pre-train the encoder of the recognition model by using a multilingual dataset to capture image features that are not language specific. Our method also pre-trains the decoder of the recognition model by using the resource-poor language's dataset to ensure its suitability for the resource-poor language. This achieves accurate recognition in the resource-poor language even though the dataset for the resource-poor language is limited. Japanese scene text recognition experiments using small publicly available Japanese dataset confirmed the improvement in the recognition accuracy offered by the proposed method.


\bibliographystyle{ACM-Reference-Format}
\bibliography{refs}










\end{document}